联系电话：

# 基于事实信息核查的虚假新闻检测综述

杨昱洲[1,2]，周杨铭[1,2]，应祺超[1,2]，钱振兴[1,2]，曾丹[3]，刘亮[4]

（1. 复旦大学计算机科学技术学院，上海 200433；
2. 复旦大学文旅部数字文化保护与旅游数据智能计算重点实验室，上海 200433；
3. 上海大学通信与信息工程学院，上海 200444；
4. 中央广播电视总台视听新媒体中心技术应用部，北京 100026）

**摘要：**基于深度学习的虚假新闻检测领域内已有许多开创性的方法能通过特征提取与检测的方式进行自动检测假新闻的任务，通常使用预训练模型提取新闻内容的特征，并开发算法使用这些特征进行检测。许多此类方法通过找到假新闻中通行的特征模式（例如写作风格、常用词等）来判别假新闻。但模型的高性能严重依赖于大量高质量标注数据的训练。然而在实际应用场景中，不仅获取、标注数据十分困难，新伪造的虚假新闻往往还会避免采用以往假新闻的写作风格，导致了模型在时间性上缺乏泛化能力。近年来事实核查在虚假新闻检测领域的发展为解决上述问题提供了新的研究思路，基于事实信息的虚假新闻检测提供了更可靠的检测解释性，通过对事件的真实性、描述与事实的匹配程度等的查验，很大程度上突破了以往方法依赖文本风格特征所带来的检测偏置。本文从任务和问题、算法策略、数据集等角度出发，对当前基于事实信息的虚假新闻的研究成果进行梳理和总结。首先，本文系统性地阐述了基于事实信息的虚假新闻检测的任务定义与核心问题。其次，从算法原理出发，对现有的检测方法进行归纳总结。之后，对领域内的经典与新提出的数据集进行了分析，对各数据集上的实验结果进行了总结。最后，本文概括性地阐述了现有方法的优势和劣势，提出了几个该领域方法可能面临的挑战，并对下一阶段的研究进行展望，期望为领域内的后续工作提供参考。

# A Survey on Fake News Detection Based on Fact Verification

YANG Yuzhou[1,2], ZHOU Yangming[1,2], YING Qichao[1,2],

QIAN Zhenxing[1,2], ZENG Dan[3], LIU Liang[4]

(1. School of Computer Science, Fudan University, Shanghai 200433, China;
2. Key Laboratory of Culture & Tourism Intelligent Computing of Ministry of Culture & Tourism, Fudan University, Shanghai 200433, China;

3. School of Communication & Information Engineering, Shanghai University, Shanghai 200444, China;
4. Technical Application Department of Audiovisual New Media Center, China Media Group, Beijing 100026, China.)

**Abstract:** There are many groundbreaking methods in the field of fake news detection based on deep learning that can automatically detect fake news through feature extraction and detection. A common methodology framework consists of extracting features from news content by pre-trained models and developing algorithms for detection. Major approaches within the scope identify fake news by learning common feature patterns in them, such as writing style, word usage, etc.. The performance of such models highly relies on large well-annotated data sets, but obtaining and annotating fake news data is laborious. Moreover, newly forged fake news often avoids utilizing the writing style of previous fake news, resulting in poor generalization ability in terms of timeliness. In recent research, fake news detection based on fact verification provides new ideas to address the above problems. Approaches within the scope verify the authenticity of the news event, matching between description and factual information, and so on, to provide more reliable and explicable detection, greatly addressing the bias of previous methods that rely on semantic and writing style features. In this paper, we sort out the research findings of fact-based fake news detection from the perspectives of tasks and problems, algorithms, datasets, and so on. First, this paper illustrates the task definition and core problems of fact-based fake news detection. Next, existing approaches are summarized and organized in terms of algorithms. Subsequently, classic and newly published datasets are analyzed, and extended by summarizing experimental results evaluation. Finally, this paper elaborates on the pros and cons of existing approaches from an overall perspective, pointing out some substantial challenges and expectations of this field. It is expected to provide references for future works in the field of fake news detection.

## 1 引言

互联网的广泛运用与其方便快捷的特性改变了本时代的信息获取和消费方式。然而，这也为假新闻的传播提供了便利，虚假新闻可以通过社交媒体和其他网络媒介迅速获得关注与传播。在其防护与治理方面，互联网上的海量媒体内容使得人工查验虚假新闻变得非常棘手，这不仅增加了社交平台的运营成本，也不利于在虚假新闻传播的早期进行阻止。得益于大规模数据集和深度学习技术的发展，基于深度学习的自动虚假新闻检测模型获得了科研工作者的广泛关注[1-3]。

虚假新闻检测任务是一个包含多种技术的分类任务。目前，国内外主流的基于深度学习的虚假新闻检测方法大致包含三步，首先是提取单一或多模态（包括文本、图像、视频等）深度学习特征，随后开发独特的算法对这些特征进行处理，最后通过二元分类来识别假新闻。现有许多热门方法可以被归类于“基于内容模式的检测方法”，此类方法使用预训练模型提取特征，然后配合以其他深度学习组件，

基金项目：国家自然科学基金联合基金项目(U20B2051)
第一作者：杨昱洲（2000-），男，博士研究生，主要从事多媒体安全研究。Email: yangyz22@m.fudan.edu.cn
通讯作者：钱振兴（1981-），男，博士，教授，主要从事多媒体安全、人工智能安全、文旅智能计算等方面的研究。Email: zxqian@fudan.edu.cn

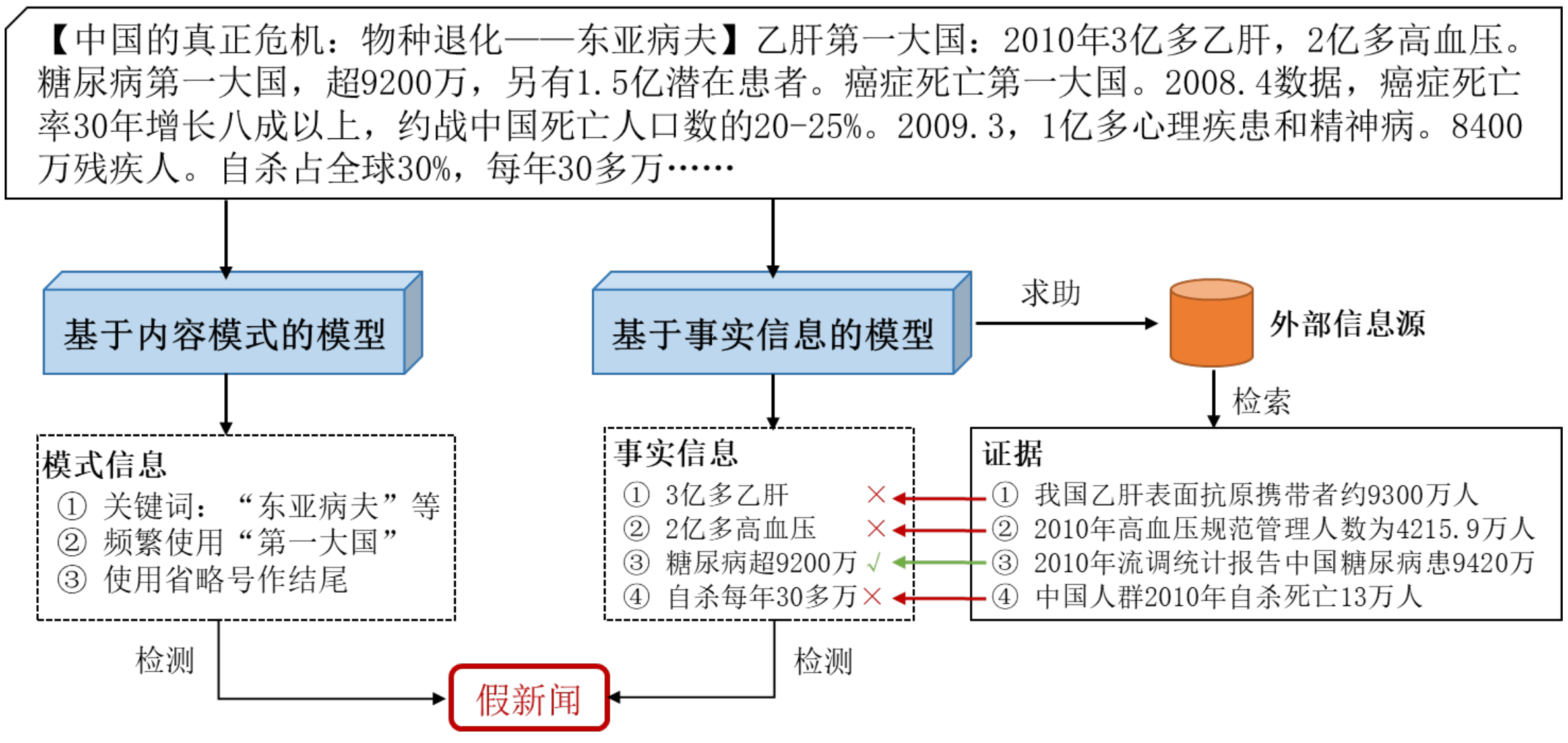

**图 1 基于内容模式的模型和基于事实信息的模型的工作流程**

尝试使模型学习到训练集中真新闻与假新闻各自的一般模式特点，从而在测试阶段通过判断测试集新闻的模式与哪种更相符以进行分类。以仅使用新闻文本的方法为例，基于内容模式的检测方法尝试找到虚假新闻中所共有的模式，例如语言特征[4, 5]、写作风格[6, 7]、情绪模式[8-10]等。由于模型所接触到的数据总是有限的，此类方法面临过拟合的问题。解决方法是为数据集引入更多的具有分布特征的信息，例如引入新闻的元数据[11]、新闻图像[12]、新闻视频[13]等等。尽管这使此类方法可正确检测的新闻不断增多，但基于内容模式的检测方法始终有其无法克服的挑战。

由于基于内容模式的模型学习到的知识是基于对训练数据集的归纳，导致模型面临如下几个挑战。首先，训练数据集的分布特征往往不具有全局代表性。其表现为各网站所发布的新闻往往具有不同模式特征，导致在某一网站上表现良好的检测模型在另一网站的新闻上检测效果欠佳。其次，由于新闻信息往往具有较强的时效性，导致模型所学习到的知识不能完全泛化到新出现的新闻上[14]。第三，大语言模型（Large Language Model，简称LLM）的出现使得生成假新闻与改写人造假新闻变得更容易[15]，这使得要通过文字模式来鉴别假新闻更加困难。为应对以上挑战，一个研究思路为使用“基于事实信息的虚假新闻检测”方法，此类方法通过对新闻所描述的各事实信息，例如事件时间、地点、具体数据等，与已知的证据进行比对查验来辨别假新闻。与基于内容模式的检测不同，基于事实信息的检测往往不受写作风格、惯用词汇等的影响。为进一步阐述基于事实信息的模型与基于内容模式的模型的区别，如图1所示，对于同一条新闻，基于内容模式的模型倾向于关注语义与词汇，判断它们是否符合虚假新闻的模式。而基于事实信息的模型则从外部信息源检索与新闻内容相关的证据，根据证据对新闻中事实信息的支持与否进行综合判断并检测。鉴于二者检测时的根据信息不同，可知基于事实信息的模型不容易受语义、词汇等信息带来的偏置的影响。

基于事实信息的虚假新闻检测方法在流程上通常更加复杂，其工作流程基本符合事实核查（Fact-checking）任务工作流程的划分[16]，即包括了待核查内容检测、证据检索、事实验证三个部分。现有的基于事实信息的虚假新闻检测工作也围绕这三个环节展开研究。对待检测内容检测的研究，其目的是找到新闻文本中需要进行证据查验的内容，也就是找到其中的事实信息。如果能准确定位到待查验内容，则证据收集会更准确，检测效果会更好。一系列工作针对该流程进行了研究[2, 17-20]。证据检索步骤的目的是找到高质量且相关的证据以提升检测效果，现有研究使用搜索引擎[21-23]、相似度算法[24, 25]、维基知识图谱[26, 27]等方法对证据进行检索。一些方法还使用自动算法来生成证据对模型进行训练[28-30]。事实验证步骤根据证据所提供的信息对待查验内容进行查证，各深度学习方法被应用于该步骤，例如图神经网络[27, 31]、实体链接[26]、注意力模块[22]、LLM[20]等。除专门针对以上目标进行的研究之外，许多工作围绕不同的目标展开研究，包括检测可解释性[32, 33]、小样本学习能力[34]等。各种新理论、新方法的研究不断出现，使虚假新闻检测领域内缺少一篇综述

对基于事实信息的虚假新闻检测的相关工作进行梳理总结的综述。

本文全面地梳理了虚假新闻领域内近年来符合“基于事实信息的检测”这一思路或目标的相关工作。在对以往经典文献进行归纳的同时，侧重梳理了近几年发表于各国际学术会议、期刊的前沿工作，并进一步讨论了事实核查如何与基于事实信息的虚假新闻检测结合，以及在此基础上为后续研究者提供具有潜力的研究方向。

**表 1 数据集对比**

| 数据集 | 待检测样本数 | 标注类别 | 证据 | | |
|---|---|---|---|---|---|
| | | | 类型 | 来源 | 人工标注 |
| FEVER[24] | 185,445 | 3 | 文本 | 维基 | 是 |
| HOVER[35] | 26,171 | 3 | 文本 | 维基 | 是 |
| TabFact[36] | 92,283 | 2 | 表格 | 维基 | 是 |
| InfoTabs[37] | 23,738 | 3 | 表格 | 维基 | 是 |
| FEVEROUS[38] | 87,026 | 3 | 文本/表格 | 维基 | 是 |
| FACTIFY-5WQA[20] | 391,041 | 3 | 文本 | 多来源 | 是 |
| FactKG[26] | 108,674 | 2 | 知识图谱 | WebNLG[39] | 是 |
| PolitiFact[40] | 3,568 | 6 | 元数据/文本 | 网站 | 否 |
| LIAR[41] | 12,836 | 5 | 元数据 | 网站 | 否 |
| Verify[42] | 422 | 2 | 文本 | 搜索引擎 | 否 |
| MultiFC[23] | 36,534 | 2-40 | 元数据/文本 | 搜索引擎 | 否 |
| Snopes[21] | 4,341 | 2 | 文本 | 网站 | 否 |
| RAWFC[33] | 2,012 | 6 | 文本 | 网站 | 否 |
| LIAR-RAW [33] | 12,590 | 6 | 文本 | 网站 | 否 |
| X-Fact[43] | 31,189 | 7 | 元数据/文本 | 搜索引擎 | 否 |
| MOCHEG[25] | 18,583 | 3 | 文本/图像 | 网站 | 是 |
| CHEF[44] | 10,000 | 3 | 元数据/文本 | 搜索引擎 | 是 |
| $MR^2$ [45] | 14,700 | 3 | 文本/图像 | 搜索引擎 | 否 |
| FAVIQ[46] | 188,376 | 2 | 文本 | 维基 | 否 |
| COVIDLies[47] | 6,761 | 3 | 文本 | 维基 | 是 |
| COVID-Fact [48] | 4,086 | 2 | 文本 | 网站 | 否 |
| Check-COVID [49] | 1,504 | 3 | 文本 | 期刊 | 是 |
| NewsCLIPings[50] | 85,360 | 2 | 文本/图像 | 搜索引擎 | 否 |

# 2 相关工作

## 2.1 数据集

本文研究了基于事实核查的假新闻检测方法。近年来，许多高质量的假新闻数据集提出了带有证据的方法。其中，Snopes[21]和PolitiFact[40]通过向Microsoft Bing API发送每条待检测文本作为查询的方式，检索相关的证据文章，并通过过滤与Snopes和PolitiFact网站相关的文章，并计算相关性分数来确定证据使用。LIAR数据集[41]包含了12,836条从PolitiFact网站上摘取的内容，包含了经济、医疗、政治等多领域内容，并人工为每一条内容进行了多达六种可能真实性的标注。一系列工作针对LIAR数据集进行了补充，例如LIAR-PLUS[51]、LIAR-RAW[33]等。Augenstein等人[23]考虑到各网站之间数据分布的差异，因此提出了一个含有收集自各网站的假新闻的数据集MultiFC。FEVER[24]是一个与现实较为贴近的证据查验数据集，其中的各条待查验文本需要对应的一篇维基百科文章进行查验。HoVer[35]包含需要使用来自维基百科

的文章进行查验的内容，相比FEVER，HOVER要求模型找出更多相关文章进行查验，要求模型具有较强的推理能力。FEVEROUS[38]同样是由来自维基百科的证据构建的数据集，包含87,026条待查验文本与匹配的证据，该数据集还包含了表格这样的结构化数据。FAVIQ[46]由188k条来自现有问题语料库的文本组成，是一个由问答任务改造的数据集。Rani等人[20]结合FEVER、HoVer等多个现有数据集提出了FACTIFY-5WQA数据集。

一些其他工作针对特定的研究目标发布了特制的数据集。过去几年，为应对新冠病毒（Covid-19）大流行，一些工作专门针对相关谣言的自动辟谣进行研究并发布了相关数据集，包括COVIDLIES[47]、COVID-Fact[48]、Check-COVID[49]。其中，Check-COVID是一个新提出的数据集，专门筛选科学依据作为证据。Chen等人[36]提出了TABFACT数据集，是一个专门针对使用表格证据进行查验的数据集，类似数据集还有InfoTabs[37]、FEVEROUS[38]。为训练模型从杂乱的证据中提取有用信息的能力，Yang等人[33]提出了RAWFC与LIAR-RAW。Fajcik等人[18]在FEVER数据集的基础上标注了一个高亮了证据中最有效内容的数据集TLR-FEVER以训练模型进行词汇级查验。Yao等人[25]提出了一个有关键证据标注的多模态数据集MOCHEG。Kim等人[26]发布FactKG数据集，这是一个专门面向基于知识图谱的事实核查提出的数据集，包含多种推理关系。Abdelnabi等人[50]使用Google APIs构建了一个多模态数据集，包含了文本证据与图片证据。Hu等人[45]发布了MR2数据集，该数据集包含多模态多语言样本，使用Google API进行构建。此外，一系列含有多语言的数据集也吸引了研究者的关注，包括Verify[42]、XFact[43]。

虚假新闻检测领域内虽然有一系列中文数据集，包括Weibo-16[2]、WeChat[52]、Weibo-20[10]、Weibo-21[11]，但这些数据集主要面向基于模式的单模态或多模态识别。Hu等人[44]提出的CHEF数据集是第一个中文的专门面向基于事实信息与证据的虚假新闻检测数据集，该数据集包含收集自Piyao、TFC、Cnews网站的一万条待检测文本与匹配的证据，并对证据进行了人工筛查与标注。

本文将以上数据集工作归纳整理在表1中。表头中，标注类别代表了样本标签类别数，具体标签类别视具体工作而定，例如，FEVER工作中按证据与待检测文本Support/Refute/NEI（Not Enough Information）关系进行了标注，而PolitiFact工作则对待检测文本进行了六个标签的标注。本工作整理了证据类型、来源与是否对证据进行了人工标注，人工标注通常是声明证据相关性、标注关键证据等，具体处理方式根据工作而定。

## 2.2 检测方法

针对以上数据集，研究人员提出了许多专门用于这些数据集的假新闻检测网络。较早期的基于事实核查虚假新闻检测工作是基于LIAR数据集展开的，其中，Alhindi等人[51]发现，无论基于什么样的机器学习模型（使用轻量级的支持向量机或较为复杂的双向LSTM），在使用目标新闻关联证据作为模型的额外输入时，分类网络的整体表现都会比单纯使用目标网络作为输入时要好，该结论也引发了一定的反响，提升了基于事实核查虚假检测研究的研究热度。此后，许多工作使用其他数据集或网络架构进行性能优化。Atanasova等人[53]使用DistilBERT网络架构进行新闻与事实内容的联合学习，不仅预测新闻真实性，还利用Transformer的Decoder结构进一步生成新闻判决结果的解释性语句。Popat[22]等人基于双向LSTM 结构与PolitiFact数据集，可以自动将原始新闻及其相关的证据性新闻进行特征提取与融合，从而给出新闻真实性判断。Ma等人[54]使用GRU网络与基于一致性的注意力模块建模证据中的句子对待检测文本的影响。Vo等人[55]进一步提出多层级多尺度注意力网络（简称MAC），使用分层多头注意力网络循序渐进地结合目标新闻与证据的词向量特征，并额外考虑了新闻与证据来源对真实性的影响，将新闻发布者与出版社的特征信息也融入到特征处理过程中。不过上述方法仅支持处理文本形式的新闻，不支持图像作为输入。Xu等人[31]认为，使用以LSTM为代表的时序模型可能会导致新闻特征提取时无法有效地将句子中长距离关键性词语联系起来，网络也因此欠缺对虚假新闻检测核心特征抽取与无效特征过滤的能力。为此，作者应用图神经网络来建模新闻和证据中的长距离语义关系，首先根据新闻与证据内容分别构建关系图，再使用图网络相邻节点信息传播更新各节点特征，并根据每个节点计算得到的重要性分数保留前$k$个重要节点，最后，将剩余节点特征根据注意力机制进行特征加权，得到可送入多层感知机的一维向量，从而输出最终的新闻虚假程度判决结果。

上述方法直接将检索到的信息作为证据来辅助虚假新闻检测，而针对如何检索与预处理每条待检测新闻所对应的证据，许多研究提出了有效且具有启发性的方法。Nie等人[56]在基于FEVER数据集提出了一个集合了证据文章检索、内容语句选择与新闻真假判决三个功能的模型，其中，证据文章检索模块可以根据目标新闻的文本特征直接找寻Wikipedia库内相关的证据性内容，而不依赖于任何知识先验或人工介入，从而能够使事实核查虚假新闻检测更高效、更智能。Wu等人[32]指出，所收集得到的证据可能存在质量方差，也即可能因情感或观念倾向存在错误信息，干扰正确判断。因此，作者团队提出三项工作来提升证据查验过程的可解释性与可操作性。在文献[32]中，作者首先提取所有证据的平均全局特征，并将其与证据特征进行逐条比对，抑制与全局特征有显著区别的证据，提升证据的质量与可靠性，结果证明此方法可以弱化单一证据文章对检测结果的影响。在文献[57]中，作者进一步设计了局部交互模块与全局交互模块，捕捉目标新闻整体的语义信息以及证据中的相关核心观点，接着，分析目标新闻与证据之间的语义差异与矛盾，最后将上述过程得到的细化特征处理为最终的判决结果。在文献[58]中，作者设计内部-交互结构，进一步在目标信息与证据特征提取过程中增强信息交互过程，从而使得信息提取过程中能够结合所有提供的信息，去除无效冗余，使网络关注于对于真假判断具有决定性的部分。以上方法在Snopes与PolitiFact数据集上取得了较好的效果。Yang等人[33]则认为许多有待检测的新闻可能在传播初期无法及时有效找寻到具有强关联性的佐证新闻，可能导致判决准确性下降，该文献提出CoFED，基于RAWFC与LIAR-RAW两个数据集，使用未经验证的证据做假新闻检测，通过收集各方观点为检测的可解释性提供支持，首先提取观点语义特征，其次计算各观点与目标新闻相关性，保留最关键观点，最后，以多文本摘要（Multi-document Extractive Summarization）的形式基于最关键观点为判决结果生成可解释性文本。

在一些最近的工作中，Claim-Dissector[18]将证据文章进行切分，查验每一部分证据与原文的支持关系，在最终检测时线性地集合所有部分的结果，从而允许模型量化每一部分证据的利害与作用。该方法在多个数据集上展现了良好效果。Xu等人[59]认为现有方法在目前的数据集上在学习事实信息查验的同时容易受到待检测文本与证据的模式信息的影响，因此提出了一个方法来处理模型推理阶段时遇到的模式信息，以让模型专注于对事实信息的处理。Rani等人[20]提出了一个5W框架（who、what、when、where、why）进行基于问题-回答模式的事实核查，并为数据集FACTIFY-5WQA按照该框架的需求进行了标注。经实验验证，5W框架一定程度上有利于基于证据的检测。FEVEROUS数据集吸引了广泛的研究兴趣[60-62]。Hu等人[62]提出的DCUF执行自然语言与表格的双通道并行检验。之后的UNIFEE[63]使用图网络联合对自然语言和表格信息进行建模，允许两种信息在流程当中执行交互。

Yao等人[25]使用MOCHEG实现了一个检索、查验、解释、判断俱备的端到端网络进行多模态的基于证据查验。CCN[50]使用其构建的大规模多模态数据集，充分利用图像标题和文本来获取大量网络在线证据并检测新闻与证据之间（文本-文本和图像-图像的一致性，以达到快速检测多模态新闻的目的。

近段时间以来，T5[64]、GPT[65]等生成式LLMs被应用于该领域的工作。Pan等人[19]提出将查验复杂文本的过程分解为多个步骤的想法，并提出了一个模型ProgramFC。该模型使用LLMs来生成推理路径，在FEVER与HOVER两个数据集上的测试表现了模型在处理推理深度加深方面的优越性与客观的可解释性。Zeng等人[66]研究了零样本与小样本情况下使用LLMs进行证据查验的问题，提出了ProToCo方法，该方法生成待检测文本的不同变体，通过衡量证据与各变体的语义一致性实现对LLMs的小样本微调。在针对中文数据集的研究方面，Sheng等人[67]为Weibo-20从辟谣网站收集证据并提出了Pref-FEND模型，试图通过图卷积神经网络与注意力模块实现“基于模式信息”与“基于事实信息”的联合检测。Hu等人[68]面向证据的质量与验证过程展开研究，提出了一个包含检索、查询、验证三个步骤的方法，该方法通过综合考量证据的可信度与说服力执行可解释的查验，并在CHEF数据集上验证了其有效性。Zhang等人[69]提出一项利用预训练语言大模型先验知识与少样本（Few-shot）提示词的具有可解释性虚假新闻检测框架，首先，向大模型提供虚假新闻检测若干样例，接着，引导大语言模型将目标新闻拆解为若干子问题，并利

用大模型对自然语言的分析能力对每个子问题的核心含义用疑问句的形式进行具体化，最后利用大模型支持的在线查询功能对每个子问题进行核查，并得到最终判决结果。该方法不仅在RAWFC与LIAR数据集上得到了很好的虚假新闻检测准确率与可解释性结果，也提供了新颖的利用预训练大模型进行事实核查假新闻检测的新思路。

## 3 总结与展望

基于事实信息的虚假新闻检测在一系列研究工作中展现出了较可靠的检测效果，具有很大的应用价值。目前，专门面向基于事实信息的虚假新闻检测的文献综述相对缺失，且该领域不同程度地与知识问答、事实核查等领域有所交集，不利于科研工作者对该领域进行快速且深入的了解。为应对这一困难，本文对目前基于事实信息的虚假新闻检测工作以及涉及领域内的相关工作根据数据集的使用情况进行了分类概括与阐述。所介绍的方法包括了证据检索、证据查验、查验解释等不同研究目的的基于深度学习的方法。重点梳理了经典与新提出的方法和数据集。

总体而言，基于事实信息的虚假新闻检测取得了很大的进展，但尚存在一些局限于需要克服的挑战。首先，各类方法由于数据集泛化性的限制，都存在其各自使用的场景和局限性。一些方法尝试通过优化数据集解决这个问题[20, 44]，但缺少从方法上的根本解决方案。其次，在多步推理上尚缺乏足够多的研究，导致现有方法往往无法充分利用证据。一些具有较高推理难度的数据集[35, 45]为该方向提供了研究基础。

基于以上分析，基于事实信息的虚假新闻检测任务可能存在一下几个值得深入研究的方向。一是针对方法通用性或泛化性的研究，包括使方法可以应用于更多的场景、提高方法对分布外数据的泛化能力等。例如，Yue等人[34]尝试使用元学习技术让模型学习到检测不同领域新闻的能力，Zeng等人[66]研究在测试集上模型零样本学习的方法以提高模型的适应性。Liu等人[70]使用对抗学习解决分布外数据样本的问题。二是结合事实核查、知识问答、知识图谱等相关领域的技术服务于虚假新闻检测。前文所提及的一系列工作，包括FactKG[26]、FACTIFY-5WQA[20]等对自然语言的特点展开研究，为未来工作更精细地考虑使用事实信息提供了思路。三是利用大语言模型进行推理过程的研究。大语言模型允许算法模型具有更多的可变性，近年来，大模型通过链式思考策略获得了较强的推理能力（Chain of Thoughts，CoT）[71]，一些工作使用该策略在取得了不错的推理效果[72, 73]。基于事实信息的虚假新闻检测在未来有良好的发展空间，尤其是大语言模型的广泛使用更易导致虚假信息的传播，本文希望为未来的基于事实信息的虚假新闻检测工作提供参考。

## 参考文献（References）：